\documentclass[sigconf, authorversion, nonacm]{acmart}
\usepackage[font=small,skip=0pt]{caption} 
\usepackage{lipsum} 
\usepackage[capitalise, noabbrev]{cleveref}
\usepackage{array}
\usepackage{multirow}

\AtBeginDocument{%
  \providecommand\BibTeX{{%
    \normalfont B\kern-0.5em{\scshape i\kern-0.25em b}\kern-0.8em\TeX}}}

\setcopyright{acmcopyright}
\copyrightyear{2018}
\acmYear{2018}
\acmDOI{XXXXXXX.XXXXXXX}

\acmConference[Conference acronym 'XX]{Make sure to enter the correct
  conference title from your rights confirmation emai}{June 03--05,
  2018}{Woodstock, NY}
\acmBooktitle{Woodstock '18: ACM Symposium on Neural Gaze Detection,
 June 03--05, 2018, Woodstock, NY} 
\acmPrice{15.00}
\acmISBN{978-1-4503-XXXX-X/18/06}

\acmSubmissionID{1102}

\begin{document}

\title[CaptAinGlove]{CaptAinGlove: Capacitive and Inertial Fusion-Based Glove for Real-Time on Edge Hand Gesture Recognition for Drone Control}

\author{Hymalai Bello, Sungho Suh, Daniel Geißler, Lala Ray, Bo Zhou and Paul Lukowicz}
\affiliation{%
   \institution{DFKI and RPTU Kaiserslautern-Landau}
   \city{Kaiserslautern}
   \country{Germany}}

\renewcommand{\shortauthors}{Bello, et al.}

\begin{abstract}
We present CaptAinGlove, a textile-based, low-power ($\leq 1.15 Watts$), privacy-conscious, real-time on-the-edge (RTE) glove-based solution with a tiny memory footprint ($\le 2MB$), designed to recognize hand gestures used for drone control.
We employ lightweight convolutional neural networks as the backbone models and a hierarchical multimodal fusion to reduce power consumption and improve accuracy.
The system yields an F1-score of 80\% for the offline evaluation of nine classes; eight hand gesture commands and null activity.
For the RTE, we obtained an F1-score of 67\% (one user).
\end{abstract}




\received{20 February 2007}
\received[revised]{12 March 2009}
\received[accepted]{5 June 2009}

\maketitle

\vspace{-15pt}
\section{Introduction}
\label{sec:Intro}
Human-robot interaction (HRI) has emerged as a significant field that focuses on optimizing the interaction between users and robots by designing interfaces that meet users' needs.
HRI is relevant in the smart factory because it improves efficiency and safety \cite{HumanRobot}.
Besides, it can empower workers with human-centered artificial intelligence \cite{TowardIndustry5}.
A particular application of HRI is drone control. 
Drones have revolutionized smart factories by covering a wide range of applications, including acting as infrastructure managers for rapid communication on large construction sites, improving safety in hazardous workplaces, and providing real-time monitoring of production processes through sensors and cameras \cite{GasLeakageDrones}.
Using drones in manufacturing has the potential to transform the industry, making it safer and more efficient \cite{DronesManufacturing}.

The default and most accurate option to operate drones are camera-based solutions\cite{Camera1,Camera2,Camera3, Camera4Deep, CameraAndIMU}.
However, concerns regarding workers' privacy and technology protection in industrial environments are critical.
In this context, alternative solutions that can provide good performance without the risk of technology leakage are highly encouraged. 
Non-camera-based wearable offers a convenient option for a privacy-aware robot control mechanism. 
The ideal wearable should be flexible and comfortable to ensure minimal disruption to the worker's schedule while maximizing efficiency.
One of the challenges the wearable community faces is bridging the gap between the performance of camera-based solutions and the accuracy of a flexible, privacy-aware wearable device. 
Using gloves for smart garments particularly interests the wearable community \cite{SmartGlove}.
Gloves are commonly used as protective equipment in various industries and offer flexibility and dexterity to generate many control patterns with the fingers and wrist. 
Textile sensors integrated into gloves provide additional advantages such as softness, comfort, lightness, and air permeability. 
Thus, a glove-based textile solution for HRI is a convenient wearable option.

We can find several textile-based sensing modalities on gloves for drone control in the literature. 
Researchers have employed various sensing technologies, including textile pressure sensors, triboelectric nanogenerators (TENG), flexible capacitive pressure sensors, piezoresistive sensors, and conductive fiber-based textile pressure sensors, among others \cite{Triboelectric, CapacitiveFlex, CapacitiveReview, Conductivefiber}. 
A common practice among state-of-the-art textile wearable glove alternatives is to use different and limited gesture dictionaries compared to camera-based solutions, mainly including going forward/backward and going to the left/right classes. 
For the case of the capacitive or textile pressure sensor options, the textile is used as a soft push button, where each textile patch is an on/off instruction. 
Textile as binary actuators (push buttons) is a simple and effective way of HRI. 
However, it lacks robustness again the null class, and the number of control instructions is limited to the number of textile patches, which is an issue if the drone/robot requires performing more sophisticated tasks \cite{DroneSophisticate}. 
For example, in \cite{Conductivefiber}, the patches are placed on the fingers tips, leading to the incorrect recognition of null activities, such as checking a smartphone, typing on a keyboard, or touching/grabbing tools, as control instructions for the drone/robot.
In \cite{CapacitiveFlex}, the authors employed a textile capacitive sensor on the back of the palm of the right/left hand, and with the other left/right hand, the user touched the patches, to generate the control signal for the drone. 
However, this solution does not account for the potential extension of gestures by including both hands in the pipeline.
In \cite{Triboelectric}, the authors used TENG to fabricate a sensing glove for gesture recognition, including sign language and drone control applications.
However, their focus was mainly on introducing the technology and its futuristic applications, thus neglecting null activities and lacking information about user experimental evaluations.

To overcome these limitations, in this paper, we introduce CaptAinGlove, a capacitive and inertial fusion-based glove-based design for real-time on-the-edge hand gesture recognition. 
CaptAinGlove incorporates textile capacitive electrodes as sensing channels on the fingers and an IMU sensor on the wrist. 
Textile capacitive sensing has demonstrated its effectiveness as a low-power consumption, cost-effective, and scalable technology for movement tracking in gesture and activity recognition \cite{bello2021mocapaci,bello2022move,zhou2023mocapose}. 
Moreover, IMU sensors have been extensively used to monitor wrist movements by researchers\cite{MMGandIMU, IMUandAudioWrist, EMGandIMUWrist}. 
CaptAinGlove is an alternative solution for hand gesture recognition that addresses critical requirements, including minimal invasiveness, low power consumption, privacy preservation, flexibility, and scalability.
While our focus is mainly on a gesture dictionary related to drone control, the same concept can be extended to hand-based HRI in various domains.
The gesture dictionary is depicted in \cref{fig:HW}\textbf{C}, which aligns with the gestures used in a camera-based solution \cite{Kiselov_2021}, using MediaPipe for real-time recognition.

The main contributions of our approach can be summarized as follows:
\begin{itemize}
    \item We present CaptAinGlove, a real-time on-the-edge solution for drone control that utilizes textile-based sensing, providing flexibility, low power consumption, and cost-effectiveness, with potential applications in sign language, gaming, and robot control. 
    \item We employ lightweight neural network models to ensure a low memory footprint, providing an embedded and sustainable solution.  
    \item We propose a hierarchical multimodal fusion to reduce power consumption and increase robustness against the null class, where the first stage detects movements and recognizes a non-null hand gesture using an inertial model. Then, using a capacitive model, the second stage recognizes the dictionary in \cref{fig:HW}\textbf{C}.
    \item Experimental results demonstrate that our approach is a step towards a wearable, textile, and privacy-friendly alternative for hand gesture recognition. 
\end{itemize}

 The paper is organized as follows; \cref{sec:Apparatus} provides detailed information on the hardware prototype, "The CaptAinGlove". \cref{sec:Fusion} introduces the details of the proposed multimodal fusion for the hand gestures recognition, and \cref{sec:Results} presents the experimental results and discussion. Finally, \cref{sec:Conclusion} concludes the paper.

\section{Apparatus}
\label{sec:Apparatus}
\begin{figure*}
    \centering
    \includegraphics[width=\textwidth]{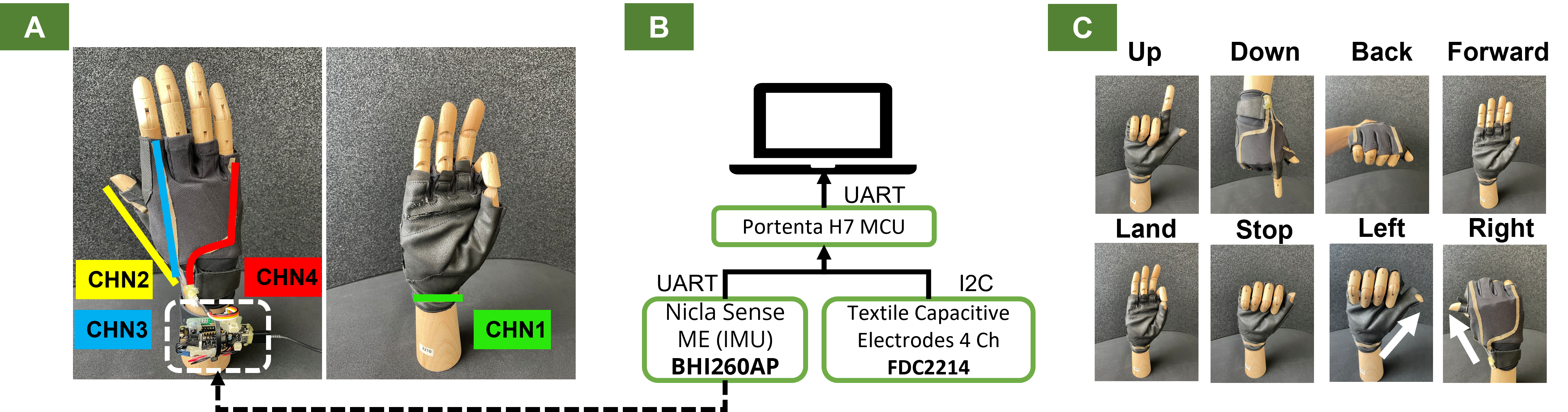}
    \caption{ CaptAinGlove Prototype; Showing the Capacitive Channels and IMU Positions on the Sports Glove \textbf{(A)}.
    Hardware Block Diagram; Sensing Connections to Main Board (Portenta H7) and PC\textbf{(B)}. 
    Hand Gestures for Drone Control Dictionary \cite{Kiselov_2021} \textbf{(C)}.     
    }
    \label{fig:HW}
    \vspace{-10pt}
\end{figure*}

\cref{fig:HW}\textbf{A} presents the prototype showing the IMU and the capacitive channels (textile electrodes) on a sports glove. 
The hardware block diagram is in  \cref{fig:HW}\textbf{B}. 
In \cref{fig:HW}\textbf{C}, the drone hand gesture dictionary is depicted. 
The dictionary comes from \cite{Kiselov_2021}. 
They presented a camera-based and real-time solution using MediaPipe. 
The gesture majority are dominant by finger patterns. 
To monitor finger movements, we proposed to use textile conductive thin patches as capacitive channels.
Moreover, the IMU-selected placement is on the wrist.
This approach reduces the number of connections, and flexibility and comfort are considered. 
Noticeably, our glove does not cover the entire area of the fingers to minimally affect the user's mobility. 

\def\oldtable{0}
\if\oldtable1
\begin{table}[t]
\caption{Apparatus Component Highlights} 
\footnotesize
\resizebox{\columnwidth}{!}{
\begin{tabular}{m{15em} m{15em}}
\hline
Component & Benefits\\
\hline 
Portenta H7
& 
\begin{itemize}
    \item Dual core STM32H747
    \item Graphics Accelerator
    \item 2MB Flash, 8MB SDRAM
    \item WiFi/BT Module
\end{itemize} \\

Nicla Sense ME
& 
\begin{itemize}
    \item BHI260AP IMU
    \item BMP390 and BME680 Pressure, Humidity, Temperature and Gas
\end{itemize} \\
FDC2214
& 
\begin{itemize}
    \item 4 Channels 28-Bit Capacitance to Digital Converter
    \item Single or Differential mode 
    \item Proximity Detection 
    \item Liquids sensing (detergent, soap, ink)
\end{itemize} \\
Textile Electrodes
& 
\begin{itemize}
    \item Shieldex Technik-tex P130+B
    \item Resistivity $\leq 2 \Omega$
    \item Double stretch direction
\end{itemize} \\
\hline
\end{tabular}}
\label{table:Components}
\vspace{-15pt}
\end{table}

\else
\begin{table}[t]
\caption{Apparatus Component Highlights} 
\footnotesize
\resizebox{\columnwidth}{!}{
\begin{tabular}{l|l}
\hline
Component & Benefits\\
\hline 
\multirow{4}{*}{Portenta H7} & \textbullet~~Dual core STM32H747    \\
                             & \textbullet~~Graphics Accelerator   \\
                             & \textbullet~~2MB Flash, 8MB SDRAM   \\
                             & \textbullet~~WiFi/BT Module         \\
\hline
\multirow{2}{*}{Nicla Sense ME} & \textbullet~~BHI260AP IMU \\
                                & \textbullet~~BMP390 and BME680, Pressure, Humidity, Temperature, and Gas\\
\hline
\multirow{4}{*}{FDC2214}    & \textbullet~~4 Channels 28-Bit Capacitance to Digital Converter  \\
                            & \textbullet~~Single or Differential mode   \\
                            & \textbullet~~Proximity Detection  \\
                            & \textbullet~~Liquids sensing (detergent, soap, ink)  \\
\hline
\multirow{4}{*}{Textile Electrodes}     & \textbullet~~Shieldex Technik-tex P130+B   \\
                                        & \textbullet~~Knit type: Stretch-Tricot    \\
                                        & \textbullet~~Resistivity $\leq 2 \Omega$   \\
                                        & \textbullet~~Double stretch direction         \\

\hline
\end{tabular}}
\label{table:Components}
\vspace{-15pt}
\end{table}
\fi

The hardware has three blocks; a main board, an inertial and environmental sensing board, and a capacitive sensing board see \cref{fig:HW}\textbf{B}. 
The main board is a Portenta H7; the main processor is the dual-core STM32H747, including a Cortex M7 running at 480 MHz and a Cortex M4 running at 240 MHz. 
Portenta H7 offers 2MB flash and 8MB SDRAM and wireless data transmissions options such as WiFi, Bluetooth classic, and BLE. 
The inertial board is a Nicla Sense with a 64 MHz ArmCortex M4 (nRF52832) and sensors such as;  IMU, air pressure, humidity, temperature, and gas. 
The capacitive board is based on the state-of-the-art capacitance to digital converter FDC2214 with four channels. 
Four capacitive channels are distributed on the glove; channel one on the wrist, channel two on the thumb, channel three on the index finger, and channel four on the little finger. 
The capacitive channels are textile electrodes based on Shieldex Technik-tex P130+B. 
The dimensions of the electrodes are 0.55 mm in thickness and between 11-15 cm long. 
The FDC2214 offers single-end and differential sensing modes. 
We use single-end mode to reduce the number of capacitive patches on the glove. 
Furthermore, the FDC2214 is configured using an external inductor of 18 uH and a capacitor of 33 pf to operate with an average frequency of 13.7 Mhz.
The sampling rate for the sensors is around 50 Hz.
The highlights of our prototype are in \cref{table:Components}.

\section{Multimodal Sensor Fusion}
\label{sec:Fusion}
As shown in \cref{fig:Results}\textbf{A}, two collaborative models were deployed for the RTE recognition of the gestures in \cref{fig:HW}\textbf{C}.  
A pre-normalization ($(x-x_{min})/(x_{max}-x_{min})$) per window is applied to the inertial and capacitive signals. 
The window size is 2s, and the window's step is 0.5s. 
The first neural network model (NN) is the inertial model with three channels as input (linear acceleration). 
This model is used to distinguish the null class from gesture detection.
The null class includes activities such as; walking and standing/sitting down, among others. 
The output of the acceleration model served as a trigger for the second model, the capacitive model.
If an activity is classified as non-null, the capacitive model is activated.
The second model fused the four capacitive channels as four independent input channels. 
The outputs of the capacitive model are the nine classes defined in the dictionary in \cref{fig:HW}\textbf{C}. 
The hierarchical approach reduces the complexity of the models, leveraging the information fusion with lightweight NNs (0.10-1.23 MB) to be deployed in tiny MCUs.
The intermediate tensor space (Arena) is 16.66 KB for the inertial model and 130.56 KB for the capacitive model.

\textbf{The Inertial Model:} 
The NN structure comprises three convolutional layers (filters=10, kernel= 10, ReLu). 
For each convolutional layer, batch normalization, max-pooling ((5,1)), and dropout (0.5) are applied.
Then it is followed by a flattening layer, a fully connected (FC) layer of 10, and an FC with softmax and two outputs. 
The training ran for 100 epochs with early stopping (patience 30 and restoring weights). 
The number of parameters of the inertial model is 2882; thus, it is a lightweight design and less susceptible to overfitting.

\textbf{The Capacitive Model:} 
The NN structure comprises two convolutional layers (filters=40, kernel= 10, ReLu). 
A normalization layer follows the first convolutional layers. 
For each convolutional layer, batch normalization, max-pooling ((5,1)), and dropout (0.3) are applied.
Then it is followed by a flattening layer, a fully connected (FC) layer of 100, and an FC with softmax and nine outputs (gesture dictionary \cref{fig:HW}\textbf{C}). 
The training ran for 200 epochs with early stopping (patience 30 and restoring weights). 
The number of parameters of the capacitive model is 49890; thus, it is a lightweight design and less susceptible to overfitting compared to a small network such as MobileNetV2 with 3.5 Million parameters. 

For both models (inertial and capacitive), the  NN optimizer is AdaDelta, with a learning rate of 0.9 and categorical cross-entropy as a loss function. 
The metric to monitor during training is accuracy. 
The NN models were trained using TensorFlow/Keras 2.12.0 framework. 

\textbf{Real-Time on the Edge:} 
TensorFlow Lite for MCU was used to generate the embedded version of the NN models. 
For RTE recognition, a sliding window scheme is employed. 
A sliding window of 2s (100 samples) with a step size of 0.5s is used as an input data frame to the NNs.
The \cref{fig:Results}\textbf{A} depicts the real-time on-the-edge procedure. 
The first step consists of movement detection (using acceleration), reducing power consumption by 10 \%.
The movement detection is based on a threshold condition ruled by $\Sigma_{n=0}^{5}=|{a_x}|_n + |{a_y}|_n +|{a_z}|_n$.
Then, the inertial model will run and detect null or gesture cases. 
In the case of $activity \neq Null$, the capacitive model will output the recognized drone control gesture in \cref{fig:HW}\textbf{C}.
The power consumption (PC) when only the movement detection is activated (only sensor data acquisition) is 0.84 W. 
Then, if a movement is detected, the inertial model is triggered, and the PC increases by 0.10 W (0.94 W). 
If the inertial model detects a gesture, the capacitive model runs, and the PC increases to 1.15 W. 
Hence, the NN models PC adds 0.31 W to the system pipeline. \footnote{USB Digital Power Meter: \url{https://www.az-delivery.de/en/products/charger-doktor}  DLA: June 02 , 2023}. 
For the RTE assessment, one volunteer performed five sessions with five repetitions of each gesture from our dictionary.         

\vspace{-5pt}
\section{Results and Discussion}
\label{sec:Results}
\begin{figure*}
    \centering  
    \includegraphics[width= \textwidth]{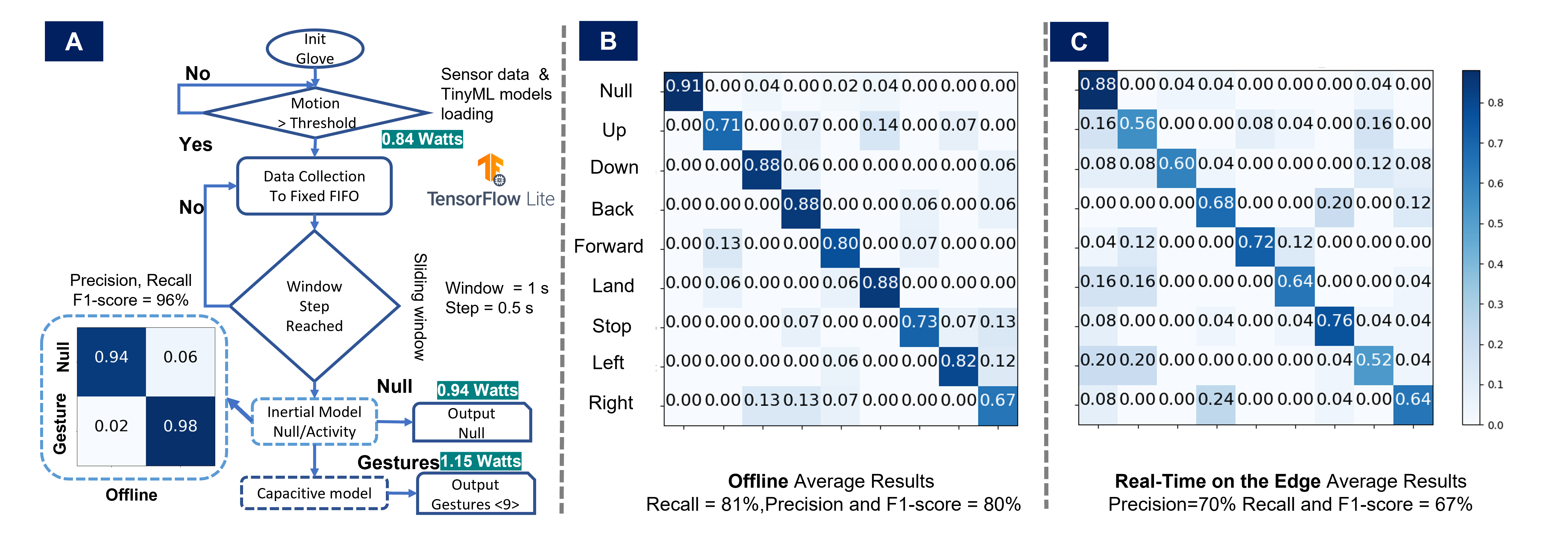}
    \caption{RTE Implementation for Hand Gesture Recognition\textbf{(A)}. 
             Results of the offline Capacitive Model; Null(0), Up(1), Down(2), Back(3), Forward(4), Land(5), Stop(6), Left(7), Right(8) and F1-score=80\%\textbf{(B)}. 
             Real-Time on the Edge Results of Capacitive Model; F1-score=67\%\textbf{(C)}.
    }
    \label{fig:Results}
    \vspace{-10pt}
\end{figure*}

\textbf{Results:}
\cref{fig:Results} shows the offline results (10 fold cross-validation) for the collaborative approach; inertial model (Null vs. Activity) in \cref{fig:Results}\textbf{A} with F1-score = 96\%  and capacitive model (gesture dictionary) with an F1-score = 80\% in \cref{fig:Results}\textbf{B}.
For the training data (offline results), one volunteer (female) participated and mimicked (randomly ten sessions) the gesture dictionary in \cref{fig:HW}\textbf{C} while wearing the system. 
The sessions were recorded on different days to ensured that our device was worn repeatedly.
The offline evaluation scheme was 10-fold cross-validation with a leaving-one-session-out. 
Each session has four random tries per gesture. 
\footnote{The participant signed an agreement following the policies of the university's committee for protecting human subjects and following the Declaration of Helsinki.}

\begin{figure}
   \centering  
    \includegraphics[width= \columnwidth]{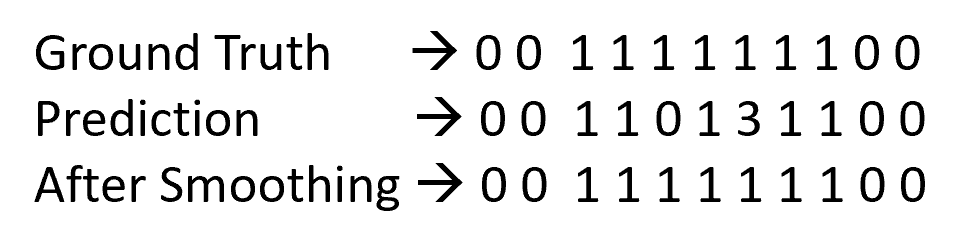}
    \caption{Example of Smoothing Temporal Windows for Continuous Recognition \cite{ContinuosRecognition}}
    \label{fig:smoothing}
    \vspace{-10pt}
\end{figure}

In \cref{fig:Results}\textbf{B} and in \cref{fig:Results}\textbf{C}, the confusion matrices for offline and online recognition are presented. 
In the offline results, we can observe confusion between the gestures, Up and Land (14\%) and Forward and Up (13\%); both pairs mainly differentiate by how the finger's upper parts move. 
The sports glove we employed does not cover the finger's upper parts to allow flexibility/comfort for the user.
There is also confusion for the case of the pairs; Stop and Right (13\%), Right and Down (13\%), Right and Back(13\%), and Left and Right (12\%). 
All these pairs have in common that the fist is closed, and their main difference is how the thumb and the index finger move. 
For specific applications such as sign language gesture recognition, the glove can be extended to cover the fingers completely to reduce confusion. 
For applications where the finger flexibility/freedom does not want to be reduced, we proposed as a future work that the inertial data (including orientation) could be fused with the capacitive information to add the wrist position in space (earth navigation frame). 
As an example of such applications, in an industrial environment, when the worker is focusing on order and picking or assembling tasks, the worker can benefit from the help of a drone/robot to ease the workflow but still needs to handle tools comfortably. 
These future solutions will positively impact the online results presented in \cref{fig:Results}\textbf{C}.  
The real-time on-the-edge results (5 sessions, re-wearing, one volunteer) in \cref{fig:Results}\textbf{C} gave an F1-score=67\%.
There was a reduction of 13\% in the F1-score between the offline and the embedded solution.
Noticeably, the RTE confusion matrix in \cref{fig:Results}\textbf{C} is based on cross-validation, shuffled, and without temporal smoothing between adjacent windows, which could improve the results in the future, as shown in \cref{fig:smoothing} . 

\textbf{Discussion:}
Our system combines inertial and capacitive sensing modalities to recognize hand gestures used for drone control using a sports glove.
The inertial information is employed as a movement detector (using a threshold). 
Later, using an inertial model,  the inertial information is used again to recognize between null and gestures from the dictionary in \cref{fig:HW}\textbf{C}. 
Then the inertial model triggers the gesture recognition with the capacitive information (nine classes).
This is similar to the approaches applied in \cite{bello2022move} and in \cite{bello2019vertical}, where the Radio Frequency Identification (RFID) signal is used as a trigger to begin gesture detection, reducing power consumption and model complexity while improving accuracy.

It is important to note that our system was tested for one participant. 
Thus more participants are still required to make it generalizable. 
The solid hardware components (MCUs and IMU) position was limited to the wrist to reduce the negative impact on the user's movements. 
The capacitive electrodes used are stretchable and soft. 
A sports glove that does not cover the entire fingers was selected to maintain the user's mobility. 
The hierarchical fusion of the inertial and capacitive information impacts the power consumption reduction by about 27\%. 
The fusion method also helps reduce model complexity and parameters to obtain lightweight neural networks to be deployed in embedded devices. 
Our approach is a step toward a textile, flexible, embedded solution for drone control. 
The main idea can be extended to other hand gesture-controlled applications where comfort, power consumption, and privacy are desirable.

On the other hand, our design has several limitations and possibilities for improvement.
The hardware prototype can be reduced in size by doing a professional encapsulated electronic design. 
The latency of the recognition in real-time is not optimized. 
The main board offers two MCUs that can work in parallel, but in our design, we allocate the entire flash (2MB) to the Cortex M7 core of the STM32H747, and the code is running only on the M7 at 480 MHz.
The latency could be improved by using the M7 to run the neural network models and the Cortex M4 (at 240 MHz) for the sensor acquisition data.  
Special attention must be given to memory access to avoid collisions and bottlenecks. 
Additionally, our work transmits the recognition results to the PC using a universal asynchronous receiver/transmitter (UART) to focus on proving the idea. 
The main board can be configured to send the recognition results by wireless communication (Bluetooth or Wifi) to improve comfort and make the system ubiquitous.
The Bluetooth/Wifi will require an external antenna and memory allocation for the wireless communication managing functions.
For the case of the RTE results, the confusion matrix is calculated based on shuffled cross-validation over fine-granular windows, which does not consider continuous sequences of windows of a single gesture where the majority of windows are true positive with sparse false detection. 
Our result could be improved by merging temporal windows from simple gap filling and event-based smoothing to selective merging based on CNN \cite{ContinuosRecognition}. 
Although temporal window smoothing techniques have been demonstrated in offline evaluations where the computation is performed on the PC, edge adaptation with resource-constrained embedded hardware is a task we will investigate in future work.

\vspace{-5pt}
\section{Conclusion}
\label{sec:Conclusion}
In this work, we have presented CaptAinGlove. 
It is a glove-based design, minimally obtrusive, low-power, privacy-friendly, flexible, and scalable solution for hand gesture recognition. 
It employs textile capacitive electrodes and inertial sensors for real-time on-the-edge recognition of hand gestures to control drones. 
Our system uses a hierarchical fusion of inertial and capacitive information to reduce power requirements and provide tiny memory models suitable for on-the-edge devices. 
In the future, this work can be extended to accommodate a variety of control-related applications, such as game control and an industrial helper tool for robot control, among others.

\section{Acknowledgments}
The research reported in this paper was partially supported by the BMBF (German Federal Ministry of Education and Research) in the projects SocialWear and STAR (Grant agreement number:
956573).


\newpage
\bibliographystyle{ACM-Reference-Format}
\bibliography{References}

\end{document}